\documentclass[runningheads]{article}

\usepackage{authblk}
\usepackage[utf8]{inputenc}
\usepackage[T1]{fontenc}
\usepackage{pslatex}
\usepackage[english]{babel}
\usepackage{fancyhdr}
\usepackage{hyperref}
\usepackage{amsmath,amssymb,amsfonts,amsthm}
\usepackage{graphicx}
\usepackage{amsmath}
\usepackage{amsthm}
\usepackage{amsfonts}
\usepackage{mathtools}
\usepackage{amssymb}
\usepackage{algorithm}
\usepackage{algorithmic}
\usepackage{microtype}
\usepackage{rotating}
\usepackage{todonotes}
\usepackage{tikz}
\usepackage{tikz-qtree}
\usepackage{xparse}
\usepackage{appendix}
\usepackage{url}
\usepackage{tabularx}
\usepackage{paralist}
\usepackage{booktabs}
\usepackage{multirow}
\usepackage{subcaption}
\usepackage{cancel}
\usetikzlibrary{calc,shapes.callouts,shapes.arrows}
\usetikzlibrary{automata}
\usetikzlibrary{positioning}
\usetikzlibrary{decorations.pathreplacing}
\usetikzlibrary{decorations.pathmorphing}
\usetikzlibrary{arrows}
\usetikzlibrary{shapes}

\newcommand{\N}{\mathbb{N}}

\newcommand{\F}{\mathbb{F}}

\newtheorem*{problem*}{Problem}

\providecommand{\keywords}[1]{\textbf{\textit{Keywords }} #1}

\begin{document}

\title{A Systematic Study on the Design of Odd-Sized Highly Nonlinear Boolean Functions via Evolutionary Algorithms}

\author[1]{Claude Carlet}
\author[2]{Marko \DH urasevic}
\author[2]{Domagoj Jakobovic}
\author[3]{Stjepan Picek}
\author[4]{Luca Mariot}

\affil[1]{{\normalsize Department of Mathematics, Universit\'{e} Paris 8, 2 rue de la libert\'{e}, 93526 Saint-Denis Cedex, France}

    {\small \texttt{claude.carlet@gmail.com}}}

\affil[2]{{\normalsize Faculty of Electrical Engineering and Computing, University of Zagreb, Unska 3, Zagreb, Croatia} \\

{\small \texttt{marko.durasevic@fer.hr, domagoj.jakobovic@fer.hr}}}

\affil[3]{{\normalsize Digital Security Group, Radboud University, Postbus 9010, 6500 GL Nijmegen, The Netherlands} \\
	
	{\small \texttt{stjepan.picek@ru.nl}}}

\affil[4]{{\normalsize Semantics, Cybersecurity and Services Group, University of Twente, 7522 NB Enschede, The Netherlands} \\
	
	{\small \texttt{l.mariot@utwente.nl}}}
	
\maketitle

\begin{abstract}
This paper focuses on the problem of evolving Boolean functions of odd sizes with high nonlinearity, a property of cryptographic relevance. Despite its simple formulation, this problem turns out to be remarkably difficult. We perform a systematic evaluation by considering three solution encodings and four problem instances, analyzing how well different types of evolutionary algorithms behave in finding a maximally nonlinear Boolean function. Our results show that genetic programming generally outperforms other evolutionary algorithms, although it falls short of the best-known results achieved by ad-hoc heuristics. Interestingly, by adding local search and restricting the space to rotation symmetric Boolean functions, we show that a genetic algorithm with the bitstring encoding manages to evolve a $9$-variable Boolean function with nonlinearity 241.
\end{abstract}


\section{Introduction}
\label{sec:intro}

Boolean functions play a fundamental role in various cryptographic applications, for example, in the design of stream ciphers based on the combiner or filter model~\cite{carlet_2021}. In this context, the cryptographic properties of a Boolean function determine whether the overall encryption system is vulnerable to certain attacks or not. Besides cryptography, Boolean functions find applications in several other closely connected domains, including combinatorics~\cite{Helleseth_Kholosha_2013}, coding theory~\cite{KERDOCK1972182}, and computational complexity theory~\cite{10.5555/1540612}. Interestingly, the same theory developed for the cryptographic properties of Boolean functions also applies to these other domains. For example, the nonlinearity of a Boolean function can be interpreted both as a cryptographic criterion and as a metric in coding theory. In the former case, the maximum nonlinearity achievable by an $n$-variable Boolean function determines the efficiency of best-affine approximation attacks and fast correlation attacks on certain stream ciphers. In the latter case, the maximum nonlinearity is equivalent to the covering radius of the first-order Reed-Muller code $RM(1, n)$, whose codewords are all linear Boolean functions of $n$ variables. As a matter of fact, the upper bound on the nonlinearity attainable by any Boolean function of $n$ variables is also called the \emph{covering radius bound}.

It is known that the covering radius bound can be satisfied with equality only when the number of variables $n$ is even; in this case, a maximally nonlinear Boolean function is also called a \emph{bent function}. On the other hand, for $n$ odd, it is still an open problem to determine a tighter upper bound on the nonlinearity. In particular, up to $n=7$ variables, this more precise bound has already been proved by means of other techniques. For a larger number of variables, the question is still open at the moment, with the available literature providing only some examples of functions that reach the best-known values. Hence, finding new functions that achieve or go beyond the best known values of nonlinearity for an odd number of variables is a relevant open problem both in the field of cryptographic Boolean functions and Reed-Muller codes.

A common approach to generating functions with a good combination of cryptographic properties is algebraic constructions, which can be divided into primary and secondary constructions. Primary constructions typically take as input the target number of variables $n$ as an input parameter, and then leverage other types of combinatorial objects (e.g., partial spreads) to construct a class of $n$-variable Boolean functions satisfying a specific set of criteria, such as optimal nonlinearity. Secondary constructions, on the other hand, take as input existing Boolean functions and generate new ones with analogous good properties, typically defined over a larger number of variables. Algebraic constructions have the advantage of a clear mathematical formulation, and they commonly work for multiple sizes, both even and odd.

On the other hand, a shortcoming of algebraic constructions is that they are not really flexible, meaning that it is difficult to employ them by optimizing for several properties of interest at once. In this respect, heuristics represent a suitable alternative for the generation of good Boolean functions. In particular, one can leverage (meta)heuristics such as evolutionary algorithms (EAs) to design functions with properties that are typically not attainable with the currently known algebraic constructions. Unfortunately, heuristics commonly struggle when considering Boolean functions of a large number of variables, due to the super-exponential growth of the search space. It has been observed in different works that genetic programming (GP) is usually able to evolve better Boolean functions than other EAs, such as genetic algorithm (GA)~\cite{00190,PICEK2016635,manzoni20}.

This paper focuses on the heuristic design of Boolean functions of an odd number of variables with high nonlinearity. As remarked above, this problem is deceptively simple in its formulation, but very hard to solve. For instance, until a few years ago, it was not known whether a $9$-variable Boolean function could achieve a nonlinearity larger than 240. While the available upper bounds theoretically allowed for its existence, no one was able to find an example of such a function for a long time. This question was positively settled in 2007 by Kavut et al. using simulated annealing~\cite{4167738}. However, the authors needed to integrate simulated annealing with specialized heuristics and limit the search space to the class of rotation symmetric Boolean functions, which is considerably smaller than the search space size for general Boolean functions (see Table~\ref{tab:searchsizes}).

The goal of this paper is to provide a systematic evaluation of various types of evolutionary algorithms applied to the optimization problem of generating highly nonlinear odd-sized Boolean functions. Since the problem is difficult due to the large size of the search space, we restrict our attention only to function sizes from $n=7$ to $n=13$ variables, thus including both an instance where the theoretical optimum is known ($n=7$) while for the remaining three ones it is still an open problem to determine it. More precisely, we evaluate three different types of genotype encodings (bitstring, symbolic, and floating-point) on these four problem instances.

This manuscript is an extended version of the paper ``A Systematic Evaluation of Evolving Highly Nonlinear Boolean Functions in Odd Sizes'' presented at EuroGP 2025~\cite{carlet2024systematicevaluationevolvinghighly}. In particular, with respect to the conference paper, here we present the following two additional contributions:
\begin{compactenum}
    \item We experiment with an additional encoding, namely a symbolic one, to evolve algebraic constructions of nonlinear Boolean functions through GP.
    \item We design two different strategies to restrict the search space of GP to rotation symmetric Boolean functions. The first strategy evolves a GP tree on a smaller number of variables such that the resulting truth table is large enough to index the set of rotation classes. The second strategy evolves a GP tree of the same target number of variables, but evaluates it only partially on the rotation class representatives.
\end{compactenum}

Overall, the findings of the conference version of the paper are confirmed: the results show that GP generally outperforms the other considered EAs in consistently evolving highly nonlinear functions. However, the optimization problem tends to be difficult, especially for the larger problem instances. In particular, we can find optimal results for certain sizes, but such solutions are rare. Already for nine inputs, none of the basic algorithms, including GP, can reach the best known value of nonlinearity achieved by the heuristic proposed in~\cite{4167738}. For this reason,
we consider two enhancement approaches. The first one adds a local search step to our GA and GP, making the results somewhat better for GA. The second one concerns the restriction of the search space to the set of rotation symmetric Boolean functions. Interestingly, with a combination of GA and local search step, we can find a rotation symmetric Boolean function with nonlinearity 241 and size 9, thereby achieving the same result of~\cite{4167738}. On the other hand, the addition of local search and rotation symmetry actually hampers the average performance of GP. We hypothesize that this result is induced by the ``superabundant'' encoding used to represent the list of rotation classes through a syntactic tree, which might actually induce GP to evolve genetic components that are not useful to solve this optimization problem.

The rest of this paper is organized as follows.
Section~\ref{sec:background} provides the necessary background information about Boolean functions, their representations, and their cryptographic properties. In Section~\ref{sec:related}, we discuss the most relevant works related to the design of nonlinear Boolean functions with (meta)heuristic optimization algorithms. Section~\ref{sec:setup} provides information about the experimental setup, while Section~\ref{sec:results} gives experimental results and concludes with a discussion of the findings.
Finally, Section~\ref{sec:conclusions} summarizes the main contributions of the paper and points out some directions for future research on this topic.

\section{Background}
\label{sec:background}

\subsection{Notation}

Let $n$ and $m$ be positive integers.
We denote the Galois (finite) field with two elements by $\mathbb{F}_{2}$. 
Moreover, we denote the Galois field with $2^n$ elements by $\mathbb{F}_{2^n}$. 
An $(n, m)$-function is a mapping $F$ from $\mathbb{F}_{2}^{n}$ to $\mathbb{F}_{2}^{m}$. 
When $m = 1$, the function is called a Boolean function (of dimension $n$) and denoted by a lowercase symbol $f$.
We endow the vector space $\mathbb{F}_2^n$ with the structure of a field, since for every $n$, there exists a field $\mathbb{F}_{2^n}$ of order $2^n$ that is an $n$-dimensional vector space. The usual inner product of $a$ and $b$ equals $a\cdot b = \bigoplus_{i=1}^{n} a_{i}b_{i}$ in $\mathbb F_{2}^n$.
Next, we discuss relevant representations and properties of Boolean functions. For more information about Boolean functions, we refer readers to~\cite{carlet_2021}.

\subsection{Boolean Function Representations}

A common way to uniquely represent a Boolean function $f$ on $\mathbb{F}_{2}^{n}$ is by using its \textbf{truth table} (TT). 
The truth table of a Boolean function $f$ is the list of pairs of function inputs in $ \mathbb F_2^n$ and function outputs, where the size of the value vector equals $2^n$. 
The output vector is the binary vector composed of all $f(x), x \in \mathbb{F}_2^n$, with a certain order selected on $\mathbb{F}_2^n$. A common option is to use a vector $(f(0),\ldots, f(1))$, which contains the function values of $f$, ordered lexicographically~\cite{carlet_2021}. 

The \textbf{Walsh-Hadamard transform} $W_{f}$ is another common representation of a Boolean function $f$. The Walsh-Hadamard transform measures the correlation between $f(x)$ and the linear functions $a\cdot x$, defined for all $a \in \mathbb{F}_2^n$ as:
\begin{equation}
W_{f} (a) = \sum\limits_{x \in \mathbb{F}_{2}^{n}} (-1)^{f(x) \oplus a\cdot x},
\end{equation}
where the sum is calculated in ${\mathbb Z}$.

\subsection{Properties and Bounds}
\label{sec:boolean_properties}

A Boolean function $f$ is \textbf{balanced} if its truth table vector is composed of an equal number of zeros and ones.
For the Walsh-Hadamard transform, a Boolean function $f$ is balanced if and only if:
\begin{equation}
W_f(0) = 0.
\end{equation}

The minimum Hamming distance between a Boolean function $f$ and all affine functions is the \textbf{nonlinearity} of $f$.
The nonlinearity $nl_{f}$ of a Boolean function $f$ can be calculated from the Walsh-Hadamard values~\cite{carlet_2021}:
\begin{equation}
\label{eq:nonlinearity}
nl_{f} = 2^{n - 1} - \frac{1}{2}\max_{a \in \mathbb{F}_{2}^{n}} |W_{f}(a)|.
\end{equation}
Since the complexity of calculating the Walsh-Hadamard transform with a naive approach equals $2^{2n}$, it is common to employ a more efficient method called the fast Walsh-Hadamard transform, where the complexity is reduced to $n2^n$. 

By the Parseval relation, it holds that $\sum_{a\in {\mathbb F}_2^n}W_f(a)^2=2^{2n}$ for any Boolean function $f$. This implies that the nonlinearity of any $n$-variable Boolean function is bounded above by the so-called covering radius bound:
\begin{equation}
\label{eq_boolean_covering}
    nl_{f} \leq 2^{n-1}-2^{\frac n 2 - 1}.
\end{equation}
Observe that Eq.~\eqref{eq_boolean_covering} cannot be tight when $n$ is odd. 
The functions whose nonlinearity equals the maximal value from Eq.~\eqref{eq_boolean_covering} are called bent. Bent functions exist only for $n$ even. 

When $n$ is odd, there is a slightly better bound that equals $2\lfloor 2^{n-2}-2^{\frac n 2 - 2}\rfloor$~\cite{568715}.
The nonlinearity $2^{n-1} - 2^{\frac{n-1}{2}}$ is called the quadratic bound\footnote{When we speak of a quadratic bound concerning general Boolean functions, this is not strictly speaking a bound but rather a value that we can try to exceed with the nonlinearity of certain functions.} since for $n$ odd, it is a tight upper bound on the nonlinearity of Boolean functions with algebraic degree at most two. This bound is also called a bent concatenation bound, as it is a tight upper bound on the nonlinearity of the concatenation of two bent functions $f$ and $g$ in $n-1$ variables.
The quadratic bound is the best nonlinearity value that can be reached for $n\leq 7$, while for $n\geq 9$, better nonlinearity values exist, see~\cite{carlet_2021}. 
We provide the best-known values for nonlinearity in Table~\ref{tab:nl}.

\begin{table}
  \centering
  \caption{Nonlinearities of Boolean functions in odd dimensions~\cite{carlet_2021}.}
  \label{tab:nl}
  \begin{tabular}{ccccc}
    &             \\
    \multicolumn{4}{c}{$n$}\\\toprule
    condition &  $7$  & $9$  & $11$  & $13$ \\ \midrule
    quadratic bound & 56 & 240 & 992 & 4032 \\\midrule
    best-known & 56 & 242 & 996 & 4040 \\\midrule
    upper bound & 58 & 244 & 1000 & 4050 \\\bottomrule
  \end{tabular}
\end{table}

\subsection{Rotation Symmetric Functions}

A Boolean function $f$ is called rotation symmetric (RS) if it is invariant under any cyclic shift of input coordinates:
$$(x_0, x_1, \ldots,  x_{n-1}) \rightarrow (x_{n-1},  x_0, x_1, \ldots, x_{n-2}).$$

The number of rotation symmetric Boolean functions is smaller than the number of Boolean functions, as the output value remains the same for certain input vectors. 
Stănică and Maitra used the Burnside lemma to deduce that the number of rotation symmetric functions is $2^{g_n}$, where $g_n$ equals~\cite{STANICA20081567}:
\begin{equation}
\label{eq:orbits}
    g_n = \frac{1}{n}\sum_{t|n}\phi(t)2^{\frac{n}{t}},
\end{equation}
and $\phi$ is the Euler totient function, which counts the number of positive integers less than $n$ that are relatively prime to it. Thus, $g_n$ represents the number of orbits, where an orbit is a rotation symmetric partition composed of vectors equivalent under rotational shifts. 
We provide the number of orbits for the rotation symmetric Boolean functions in Table~\ref{tab:rots}. Notice that for $n=9$, an exhaustive search is not practical. However, considering rotation symmetric functions does allow an exhaustive search for larger Boolean function sizes, or at least a ``simpler'' problem for heuristics. We list the search space sizes for general Boolean functions and rotation symmetric functions in Table~\ref{tab:searchsizes}.

\begin{table}
  \centering
  \scriptsize
  \caption{The number of orbits for the rotation symmetric Boolean functions.}
  \label{tab:rots}
  \begin{tabular}{ccccc}
    &             \\
    \multicolumn{4}{c}{$n$}\\\toprule
    variables&	7&	9&	11&	13 \\\midrule
    $g_n$&	20&	60&	188&	632 \\
\bottomrule
  \end{tabular}
\end{table}

\begin{table}
  \centering
  \caption{The number of Boolean functions and rotation symmetric Boolean functions.}
  \label{tab:searchsizes}
  \begin{tabular}{ccccc}
    &             \\
    \multicolumn{4}{c}{$n$}\\\toprule
    criterion & $7$  & $9$  & $11$ & $13$ \\ \midrule
    \# general  & $2^{128}$ & $2^{512}$  & $2^{2048}$  & $2^{8192}$ \\\midrule
    \# RS & $2^{20}$  & $2^{60}$  & $2^{188}$  & $2^{632}$ \\
\bottomrule
  \end{tabular}
\end{table}

\subsection{On Constructing Boolean Functions}

In the process of generating Boolean functions with specific properties, there are several options to consider.
The first deals with the type of technique used to generate Boolean functions. In general, we can use either algebraic constructions or computational techniques. Moreover, within computational techniques, we can distinguish among random search, specific heuristics, and metaheuristics.
Each of those approaches has advantages and disadvantages.
Random search represents a simple way to generate diverse functions, but it commonly struggles when the required Boolean functions are more rare (or need to fulfill multiple properties at the same time). Specific heuristic techniques are capable of obtaining Boolean functions with excellent properties, but such techniques are more difficult to develop.
Finally, metaheuristics can reach excellent results even when looking for rare Boolean functions, but do not require significant expertise to develop (except, of course, in the part of appropriate fitness function design). As such, metaheuristics could be positioned between random search and specific heuristics as a trade-off between performance and ease of development.


\section{Related Work}
\label{sec:related}

The history of using (meta)heuristics to design Boolean functions with specific properties is rather rich and spans numerous works over almost three decades~\cite{DjurasevicJMP23}.
During that time, researchers considered different Boolean functions and their properties. Some represented well-established goals like obtaining bent~\cite{FullerDM03}, or balanced and highly nonlinear functions~\cite{10.1007/BFb0028471, Picek2015}, while others considered more exotic settings like hyper-bent functions~\cite{picek18a} or quaternary bent functions~\cite{10.1007/978-3-030-16670-0_17}.
Interestingly, when considering highly nonlinear Boolean functions (but not necessarily bent ones), most works concentrate on Boolean functions in even dimensions. Next, we briefly discuss a selection of related works where we list results on Boolean functions in odd dimensions, where available.

Millan et al. were the first to apply a genetic algorithm (with hill climbing) to evolve Boolean functions with high nonlinearity~\cite{10.1007/BFb0028471}. 
Clark and Jacob used a combination of simulated annealing and hill climbing with a cost function motivated by the Parseval identity to find functions with high nonlinearity and low autocorrelation~\cite{10.1007/10718964_20}. 
The best results for 9 inputs equal 236 and 238, achieved with genetic algorithms and simulated annealing, respectively.
Burnett et al. used custom heuristics to generate Boolean functions with good cryptographic properties~\cite{burnett}. The authors reported a nonlinearity of 240 for Boolean functions with 9 inputs.

Picek et al. used genetic programming to find Boolean functions with high nonlinearity (the authors considered several fitness functions and different combinations of cryptographic properties)~\cite{10.1145/2464576.2464671}. 
Mariot and Leporati proposed using Particle Swarm Optimization to find Boolean functions with good trade-offs of cryptographic properties~\cite{10.1145/2739482.2764674}. The authors found Boolean functions in 9 inputs with nonlinearity equal to 236.

Stănică et al. used simulated annealing to search for rotation symmetric Boolean functions~\cite{StanicaMC04}. The authors constructed Boolean functions with 9 variables and nonlinearity 240.
Kavut et al. used a steepest descent-like iterative algorithm to construct highly nonlinear Boolean functions~\cite{4167738}. The authors found imbalanced 9-variable Boolean functions with a nonlinearity of 241. This represented a significant result since before it, it was unknown whether the nonlinearity could exceed 240 for Boolean functions with 9 inputs.
Kavut et al. conducted an efficient exhaustive search of rotation symmetric Boolean functions in 9 variables having nonlinearity greater than 240~\cite{Kavut2006}. They showed that there exist functions with nonlinearity 241, but there are no rotation symmetric Boolean functions with larger nonlinearity.

Kavut and Yucel used a steepest-descent-like iterative algorithm to construct imbalanced Boolean functions in 9 variables with nonlinearity $242$~\cite{KAVUT2010341}. For this result, the authors considered the generalized rotation symmetric Boolean functions class.
Carlet et al. used evolutionary algorithms to evolve rotation symmetric Boolean functions~\cite{CarletDGJMP24}. The authors reported balanced Boolean functions in 9 variables with nonlinearity 240.
Recently, Carlet et al. used evolutionary algorithms in combination with local search to obtain Boolean functions in 9 inputs with nonlinearity 241~\cite{carlet2024systematicevaluationevolvinghighly}. As far as we know, this is the first time that EAs have found such a function.

The above-discussed works construct Boolean functions directly, i.e., from scratch, making it aligned with primary construction principles. However, there are also works where metaheuristics are used to obtain secondary constructions.
Picek and Jakobovic used genetic programming to evolve secondary constructions of bent Boolean functions~\cite{evolvingconstruction}.
Mariot et al. followed the same principles with the goal of evolving hyper-bent Boolean functions~\cite{10.1007/978-3-030-16670-0_17}. Carlet et al. considered genetic programming to evolve secondary constructions of balanced, highly nonlinear Boolean functions~\cite{10.1145/3512290.3528871}. 
Mariot et al. used evolutionary algorithms to design secondary semi-bent and bent constructions of Boolean functions based on cellular automata~\cite{DBLP:journals/nc/MariotSLM22}.
None of the related works considered evolving secondary constructions to obtain highly nonlinear Boolean functions in odd sizes.

\section{Experimental Setup}
\label{sec:setup}

In this section, we present the details of the optimization algorithms considered in our systematic evaluation. We start with a description of the common characteristics shared by the algorithms, namely the encodings used to represent the genotype of the candidate solutions and the fitness function to be optimized, which targets the nonlinearity property of Boolean functions. Then, we report the remaining parameters of the algorithms that are specific to each encoding.

\subsection{Solution Encodings}
\label{subsec:sol-enc}
The search space of our optimization problem of interest is the set of $n$-variable Boolean functions $\mathcal{F}_n = \{f: \F_2^n \to \F_2\}$, with possible variations or reductions (e.g., if we restrict our attention to rotation symmetric Boolean functions). As such, the phenotype of a candidate solution is a specific mapping $f: \F_2^n \to \F_2$, which can be represented in different ways as discussed in Section~\ref{sec:background}. In the context of this article, we focus on the truth table as a phenotype representation, since from there one can easily compute the nonlinearity of a function by computing the Walsh-Hadamard transform. Hence, for any given number of variables $n \in \N$, the phenotype space is $\mathcal{P}_n = \F_2^{2^n}$, that is, the set of all $2^n$-bit vectors.

Clearly, there are various ways to define a genotype for the candidate solutions manipulated by an evolutionary algorithm, which must then be mapped to the truth table-based representation of the phenotype. In what follows, we consider three such genotype encodings: a straightforward bitstring encoding, a symbolic one based on GP syntax trees, both for directly synthesizing truth tables and for defining secondary constructions, and a floating-point encoding.

\subsubsection{Bitstring Encoding}
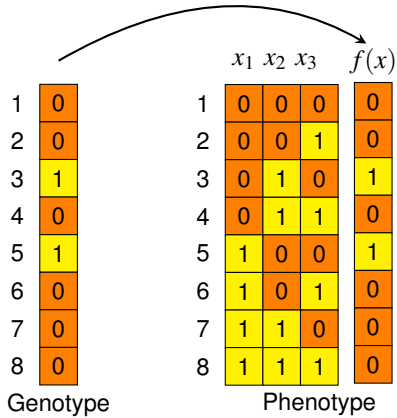
\begin{figure}
    \centering
    \begin{tikzpicture}
	[->,auto,node distance=0.8cm, empt node/.style={font=\sffamily,inner
		sep=0pt,minimum size=0pt},
	rect0 node/.style={rectangle,draw,fill=orange,font=\sffamily,minimum size=0.5cm, inner
		sep=0pt, outer sep=0pt},
	rect1 node/.style={rectangle,draw,fill=yellow,font=\sffamily,minimum size=0.5cm, inner
		sep=0pt, outer sep=0pt},
	circ1 node/.style={circle,draw,fill=red!70,font=\sffamily\bfseries,minimum size=0.7cm, inner
		sep=0pt, outer sep=0pt},
	circ2 node/.style={circle,draw,fill=blue!30,font=\sffamily\bfseries,minimum size=0.7cm, inner
		sep=0pt, outer sep=0pt},
	circ3 node/.style={circle,draw,fill=brown!80,font=\sffamily\bfseries,minimum size=0.7cm, inner
		sep=0pt, outer sep=0pt},
	circ4 node/.style={circle,draw,fill=yellow,font=\sffamily\bfseries,minimum size=0.7cm, inner
		sep=0pt, outer sep=0pt}]
	
	\node [empt node] (out)   {};
	\node [rect0 node] (t01) [below=0.3cm of out] {0};
    \node [rect0 node] (t02) [below=0cm of t01] {0};
    \node [rect1 node] (t03) [below=0cm of t02] {1};
    \node [rect0 node] (t04) [below=0cm of t03] {0};
    \node [rect1 node] (t05) [below=0cm of t04] {1};
    \node [rect0 node] (t06) [below=0cm of t05] {0};
    \node [rect0 node] (t07) [below=0cm of t06] {0};
    \node [rect0 node] (t08) [below=0cm of t07] {0};
    \node [empt node] (lab) [below=0.1cm of t08] {Genotype};
    \node [empt node] (l01) [left=0.2cm of t01] {1};
    \node [empt node] (l02) [left=0.2cm of t02] {2};
    \node [empt node] (l03) [left=0.2cm of t03] {3};
    \node [empt node] (l04) [left=0.2cm of t04] {4};
    \node [empt node] (l05) [left=0.2cm of t05] {5};
    \node [empt node] (l06) [left=0.2cm of t06] {6};
    \node [empt node] (l07) [left=0.2cm of t07] {7};
    \node [empt node] (l08) [left=0.2cm of t08] {8};
	
	\node [empt node] (x11) [right=2.3cm of out] {$x_1$};
	\node [empt node] (x12) [right=0.1cm of x11] {$x_2$};
	\node [empt node] (x13) [right=0.1cm of x12] {$x_3$};
	
	\node [rect0 node] (t11) [below=0.2cm of x11] {0};
	\node [rect0 node] (t12) [right=0cm of t11] {0};
	\node [rect0 node] (t13) [right=0cm of t12] {0};
	
	\node [rect0 node] (t21) [below=0cm of t11] {0};
	\node [rect0 node] (t22) [right=0cm of t21] {0};
	\node [rect1 node] (t23) [right=0cm of t22] {1};
	
	\node [rect0 node] (t31) [below=0cm of t21] {0};
	\node [rect1 node] (t32) [right=0cm of t31] {1};
	\node [rect0 node] (t33) [right=0cm of t32] {0};
	
	\node [rect0 node] (t41) [below=0cm of t31] {0};
	\node [rect1 node] (t42) [right=0cm of t41] {1};
	\node [rect1 node] (t43) [right=0cm of t42] {1};
	
	\node [rect1 node] (t51) [below=0cm of t41] {1};
	\node [rect0 node] (t52) [right=0cm of t51] {0};
	\node [rect0 node] (t53) [right=0cm of t52] {0};
	
	\node [rect1 node] (t61) [below=0cm of t51] {1};
	\node [rect0 node] (t62) [right=0cm of t61] {0};
	\node [rect1 node] (t63) [right=0cm of t62] {1};
	
	\node [rect1 node] (t71) [below=0cm of t61] {1};
	\node [rect1 node] (t72) [right=0cm of t71] {1};
	\node [rect0 node] (t73) [right=0cm of t72] {0};
	
	\node [rect1 node] (t81) [below=0cm of t71] {1};
	\node [rect1 node] (t82) [right=0cm of t81] {1};
	\node [rect1 node] (t83) [right=0cm of t82] {1};

    \node [empt node] (lab) [below=0.1cm of t83] {Phenotype};
    \node [empt node] (l11) [left=0.2cm of t11] {1};
    \node [empt node] (l12) [left=0.2cm of t21] {2};
    \node [empt node] (l13) [left=0.2cm of t31] {3};
    \node [empt node] (l14) [left=0.2cm of t41] {4};
    \node [empt node] (l15) [left=0.2cm of t51] {5};
    \node [empt node] (l16) [left=0.2cm of t61] {6};
    \node [empt node] (l17) [left=0.2cm of t71] {7};
    \node [empt node] (l18) [left=0.2cm of t81] {8};
	
	\node [empt node] (fx11) [right=0.4cm of x13] {$f(x)$};
	\node [rect0 node] (f000) [below=0.1cm of fx11] {0};
	\node [rect0 node] (f001) [below=0cm of f000] {0};
	\node [rect1 node] (f010) [below=0cm of f001] {1};
	\node [rect0 node] (f011) [below=0cm of f010] {0};
	\node [rect1 node] (f100) [below=0cm of f011] {1};
	\node [rect0 node] (f101) [below=0cm of f100] {0};
	\node [rect0 node] (f110) [below=0cm of f101] {0};
	\node [rect0 node] (f111) [below=0cm of f110] {0};
	
	\draw[->,thick, shorten >=3pt,shorten <=0pt,>=stealth] (out.north) [out=30, in=150] to (fx11.north);
	
\end{tikzpicture}
    \caption{Example of bitstring encoding for a $3$-variable function. The genotype is a bitstring of length $2^3=8$ used as the output column of the truth table of the function, once we sort lexicographically all vectors of $\F_2^3$.}
    \label{fig:ex-bitstring}
\end{figure}

The bitstring encoding is probably the most natural representation when optimizing Boolean functions with EAs. Indeed, since the phenotype is already the truth table of the function, one can consider the genotype as the fixed-size bitstring of size $2^n$ that encodes the output column of the table. In this case, once the order of the input vectors of $\F_2^n$ has been fixed (i.e., the lexicographic order), the transformation of the genotype into the phenotype is simply the identity mapping. Figure~\ref{fig:ex-bitstring} depicts an example of bitstring encoding for a Boolean function of $n=3$ variables.

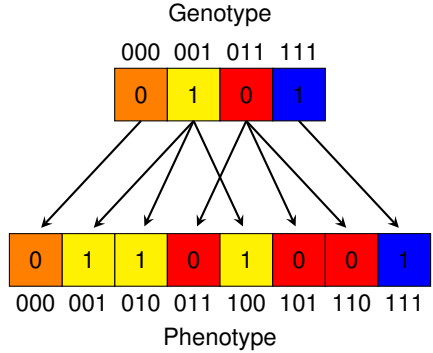
\begin{figure}
    \centering
    \begin{tikzpicture}
	[->,auto,node distance=0.8cm, empt node/.style={font=\sffamily,inner
		sep=0pt,minimum size=0pt},
	rect0 node/.style={rectangle,draw,fill=orange,font=\sffamily,minimum size=0.7cm, inner
		sep=0pt, outer sep=0pt},
	rect1 node/.style={rectangle,draw,fill=yellow,font=\sffamily,minimum size=0.7cm, inner
		sep=0pt, outer sep=0pt},
    rect2 node/.style={rectangle,draw,fill=red,font=\sffamily,minimum size=0.7cm, inner
		sep=0pt, outer sep=0pt},
    rect3 node/.style={rectangle,draw,fill=blue,font=\sffamily,minimum size=0.7cm, inner
		sep=0pt, outer sep=0pt}]
	
    \node [rect0 node] (n0) {0};
    \node [rect1 node] (n1) [right=0cm of n0] {1};
    \node [rect2 node] (n2) [right=0cm of n1] {0};
    \node [rect3 node] (n3) [right=0cm of n2] {1};

    \node [empt node] (e0) [above=0.1cm of n0] {000};
    \node [empt node] (e1) [above=0.1cm of n1] {001};
    \node [empt node] (e2) [above=0.1cm of n2] {011};
    \node [empt node] (e3) [above=0.1cm of n3] {111};

    \node [empt node] (e4) [above=0.9cm of n2.west] {Genotype};

    \node [rect1 node] (t2) [below=1.5cm of n0] {1};
    \node [rect1 node] (t1) [left=0cm of t2] {1};
    \node [rect0 node] (t0) [left=0cm of t1] {0};
    \node [rect2 node] (t3) [right=0cm of t2] {0};
    \node [rect1 node] (t4) [right=0cm of t3] {1};
    \node [rect2 node] (t5) [right=0cm of t4] {0};
    \node [rect2 node] (t6) [right=0cm of t5] {0};
    \node [rect3 node] (t7) [right=0cm of t6] {1};
    
    \node [empt node] (f0) [below=0.1cm of t0] {000};
    \node [empt node] (f1) [below=0.1cm of t1] {001};
    \node [empt node] (f2) [below=0.1cm of t2] {010};
    \node [empt node] (f3) [below=0.1cm of t3] {011};
    \node [empt node] (f4) [below=0.1cm of t4] {100};
    \node [empt node] (f5) [below=0.1cm of t5] {101};
    \node [empt node] (f6) [below=0.1cm of t6] {110};
    \node [empt node] (f7) [below=0.1cm of t7] {111};

    \node [empt node] (e8) [below=0.9cm of t3.east] {Phenotype};
    
	\draw[->,thick, shorten >=3pt,shorten <=0pt,>=stealth] (n0.south) -- (t0.north);
    \draw[->,thick, shorten >=3pt,shorten <=0pt,>=stealth] (n1.south) -- (t1.north);
    \draw[->,thick, shorten >=3pt,shorten <=0pt,>=stealth] (n1.south) -- (t2.north);
    \draw[->,thick, shorten >=3pt,shorten <=0pt,>=stealth] (n1.south) -- (t4.north);
    \draw[->,thick, shorten >=3pt,shorten <=0pt,>=stealth] (n2.south) -- (t3.north);
    \draw[->,thick, shorten >=3pt,shorten <=0pt,>=stealth] (n2.south) -- (t5.north);
    \draw[->,thick, shorten >=3pt,shorten <=0pt,>=stealth] (n2.south) -- (t6.north);
    \draw[->,thick, shorten >=3pt,shorten <=0pt,>=stealth] (n3.south) -- (t7.north);
    
\end{tikzpicture}
    \caption{Example of bitstring encoding for a $3$-variable rotation symmetric function. The genotype is a bitstring of length $g_3=4$ whose loci refer to the rotation class representatives.}
    \label{fig:ex-rotsym}
\end{figure}

Bitstring encoding can still be employed in rotation symmetric Boolean functions, although the corresponding genotype is shorter than in the generic case. Indeed, one only has to specify the output value of $f$ for each rotation class representative. The remaining input vectors obtained by cyclically shifting this representative can then be assigned to the same output value. After repeating this step for each representative, the full truth table of the function has been synthesized. Consequently, the length of the bitstring encoding genotype for a rotation symmetric Boolean function of $n$ variables is given by $g_n$, as defined in Eq.~\eqref{eq:orbits}. For each rotation class, we take the first vector in lexicographic order as a representative~\cite{StanicaMC04}. Figure~\ref{fig:ex-rotsym} depicts an example of bitstring encoding for a 3-variable rotation symmetric Boolean function. In this case, we have four rotation classes, and four lexicographically smallest representatives are respectively $000$, $001$, $011$, and $111$. The full output column of the truth table of $f$ is then expanded by copying the bit of each representative in all positions that are equivalent by cyclic shifts. Thus, the bits $000$ and $111$ are copied only in those same positions, while those of $001$ and $011$ are copied respectively in $001$, $010$, $100$ and $011$, $110$, $101$.

\subsubsection{Symbolic Encoding}
\begin{figure}
    \centering
    	\begin{tikzpicture}
	[->,auto,node distance=0.8cm, empt node/.style={font=\sffamily,inner
		sep=0pt,minimum size=0pt},
	rect0 node/.style={rectangle,draw,fill=orange,font=\sffamily,minimum size=0.5cm, inner
		sep=0pt, outer sep=0pt},
	rect1 node/.style={rectangle,draw,fill=yellow,font=\sffamily,minimum size=0.5cm, inner
		sep=0pt, outer sep=0pt},
	circ1 node/.style={circle,draw,fill=red!70,font=\sffamily\bfseries,minimum size=0.7cm, inner
		sep=0pt, outer sep=0pt},
	circ2 node/.style={circle,draw,fill=blue!30,font=\sffamily\bfseries,minimum size=0.7cm, inner
		sep=0pt, outer sep=0pt},
	circ3 node/.style={circle,draw,fill=brown!80,font=\sffamily\bfseries,minimum size=0.7cm, inner
		sep=0pt, outer sep=0pt},
	circ4 node/.style={circle,draw,fill=yellow,font=\sffamily\bfseries,minimum size=0.7cm, inner
		sep=0pt, outer sep=0pt}]
	
	\node [empt node] (out)   {};
	\node [circ1 node] (and) [below=0.8cm of out] {$\land$};
	\node [circ4 node] (xor) [below left=0.9cm and 0.2cm of and] {$\oplus$};
	\node [circ3 node] (not) [below right=0.9cm and 0.2cm of and] {$\neg$};
	\node [circ2 node] (x1)  [below left=0.9cm and 0.2cm of xor] {$x_1$};
	\node [circ2 node] (x2)  [below right=0.9cm and 0.2cm of xor] {$x_2$};
	\node [circ2 node] (x3)  [below right=0.9cm and 0.2cm of not] {$x_3$};

    \node [empt node] (e1) [below=0.2 cm of x2] {Genotype};
	
	\draw[>=stealth,thick] (x1) -- (xor);
	\draw[>=stealth,thick] (x2) -- (xor);
	\draw[>=stealth,thick] (x3) -- (not);
	\draw[>=stealth,thick] (xor) -- (and);
	\draw[>=stealth,thick] (not) -- (and);
	\draw[>=stealth,thick] (and) -- (out);
	
	\node [empt node] (x11) [right=2.3cm of out] {$x_1$};
	\node [empt node] (x12) [right=0.1cm of x11] {$x_2$};
	\node [empt node] (x13) [right=0.1cm of x12] {$x_3$};
	
	\node [rect0 node] (t11) [below=0.2cm of x11] {0};
	\node [rect0 node] (t12) [right=0cm of t11] {0};
	\node [rect0 node] (t13) [right=0cm of t12] {0};
	
	\node [rect0 node] (t21) [below=0cm of t11] {0};
	\node [rect0 node] (t22) [right=0cm of t21] {0};
	\node [rect1 node] (t23) [right=0cm of t22] {1};
	
	\node [rect0 node] (t31) [below=0cm of t21] {0};
	\node [rect1 node] (t32) [right=0cm of t31] {1};
	\node [rect0 node] (t33) [right=0cm of t32] {0};
	
	\node [rect0 node] (t41) [below=0cm of t31] {0};
	\node [rect1 node] (t42) [right=0cm of t41] {1};
	\node [rect1 node] (t43) [right=0cm of t42] {1};
	
	\node [rect1 node] (t51) [below=0cm of t41] {1};
	\node [rect0 node] (t52) [right=0cm of t51] {0};
	\node [rect0 node] (t53) [right=0cm of t52] {0};
	
	\node [rect1 node] (t61) [below=0cm of t51] {1};
	\node [rect0 node] (t62) [right=0cm of t61] {0};
	\node [rect1 node] (t63) [right=0cm of t62] {1};
	
	\node [rect1 node] (t71) [below=0cm of t61] {1};
	\node [rect1 node] (t72) [right=0cm of t71] {1};
	\node [rect0 node] (t73) [right=0cm of t72] {0};
	
	\node [rect1 node] (t81) [below=0cm of t71] {1};
	\node [rect1 node] (t82) [right=0cm of t81] {1};
	\node [rect1 node] (t83) [right=0cm of t82] {1};

    \node [empt node] (e4) [below=0.2 cm of t83] {Phenotype};

	\node [empt node] (fx11) [right=0.4cm of x13] {$f(x)$};
	\node [rect0 node] (f000) [below=0.1cm of fx11] {0};
	\node [rect0 node] (f001) [below=0cm of f000] {0};
	\node [rect1 node] (f010) [below=0cm of f001] {1};
	\node [rect0 node] (f011) [below=0cm of f010] {0};
	\node [rect1 node] (f100) [below=0cm of f011] {1};
	\node [rect0 node] (f101) [below=0cm of f100] {0};
	\node [rect0 node] (f110) [below=0cm of f101] {0};
	\node [rect0 node] (f111) [below=0cm of f110] {0};
	
	\draw[->,thick, shorten >=3pt,shorten <=0pt,>=stealth] (out.north) [out=30, in=150] to (fx11.north);
	
	\end{tikzpicture}
    \caption{Example of symbolic encoding for a $3$-variable Boolean function. The genotype is a Boolean tree where the leaf nodes correspond to the input variables $x_1$, $x_2$, and $x_3$. The output of the function is given by the root node, from which one can reconstruct the full truth table by evaluating the tree over all $2^3=8$ assignments of the leaf nodes.}
    \label{fig:ex-gptree}
\end{figure}
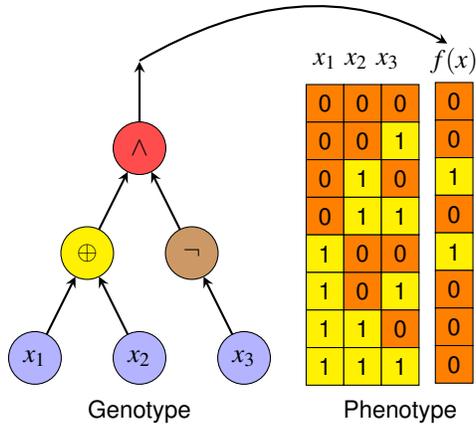

As a second genotype representation, we experimented with symbolic expressions evolved through GP. In this context, a Boolean function is encoded as a tree whose leaf nodes represent the input variables $x_1,\ldots, x_n \in \F_2$ of an $n$-variable Boolean function. The internal nodes, on the other hand, are Boolean operators that take the inputs received from their children nodes as operands, and forward the corresponding output value to their parent node. For instance, an internal node could stand for either an XOR, OR, or AND (all taking the input of two children nodes), or a NOT (taking the input from a single child node). The output of the overall function is given by the output evaluated at the root node. The genotype-to-phenotype mapping consists of evaluating the tree at all the $2^n$ possible assignments of the leaf nodes, thus synthesizing the full truth table of the function. Figure~\ref{fig:ex-gptree} provides an example of symbolic encoding for a $3$-variable Boolean function $f: \F_2^3 \to \F_2$. In particular, the symbolic expression of the function is given by: $f(x_1, x_2, x_3) = (x_1 \oplus x_2) \land (\neg x_3)$.

\begin{figure}[t]
    \centering
    \begin{subfigure}{\textwidth}
    \includegraphics[width=\textwidth]{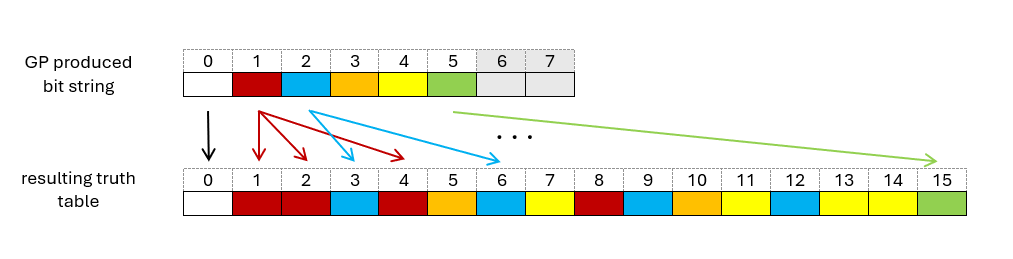}
    \caption{GP/PART}
    \label{fig:gp-part}
    \end{subfigure}
    \begin{subfigure}{\textwidth}
    \includegraphics[width=\textwidth]{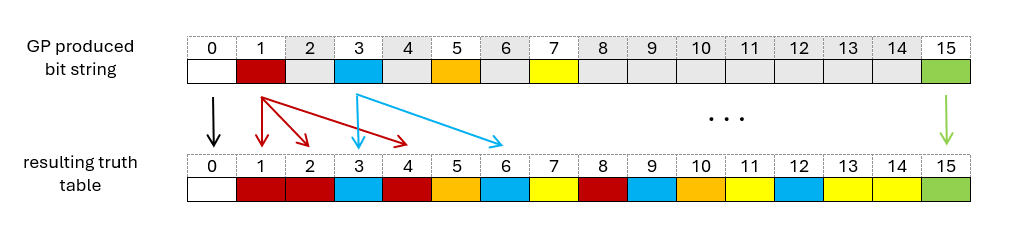}
    \caption{GP/FULL}
    \label{fig:gp-full}
    \end{subfigure}
    \caption{Example of the two approaches to represent a rotation symmetric Boolean function under the symbolic encoding (grey cells are not used).}
    \label{fig:gp-rotsym}
\end{figure}

The symbolic encoding can also be adapted to represent rotation symmetric Boolean functions, although in a less straightforward manner than for the bitstring case. In particular, given the number of variables $n \in \N$ of the target function, we explored two approaches:
\begin{itemize}
    \item We evolve a GP tree for a function of a number of variables $\tilde{n}$ so that the corresponding truth table is large enough to index all rotation class representatives of $n$ variables. In other words, we have to choose $\tilde{n}$ such that $2^{\tilde{n}} \ge g_n$. As an example, consider the situation depicted in Figure~\ref{fig:gp-part}: for $n=4$ one can evolve a Boolean tree for a function of $\tilde{n} = 3$ variables, since $2^{3} = 8 \ge g_4 = 6$. The two additional bits are then discarded. We call this approach GP/PART.
    \item We evolve a GP tree of the same size as the target function. However, we only evaluate the tree based on the assignments of the leaf nodes that correspond to the rotation class representatives, setting the other members in the class to the same value. This approach simulates what we do with the bitstring encoding for rotation symmetric functions. Figure~\ref{fig:gp-full} depicts an example for $n=4$ variables. We denote this approach as GP/FULL.
\end{itemize}

\begin{figure}
    \centering
    \includegraphics[width=0.5\textwidth]{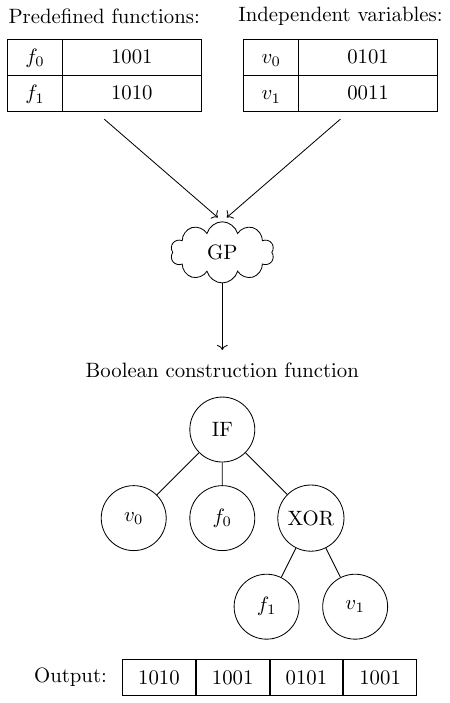}
    \caption{An example of evolving Boolean construction using $4$ seed functions of $2$ variables, with the construction resulting in a $n + 2 = 4$ variable Boolean function.}
    \label{fig:gp-sec-constr}
\end{figure}

Further, we considered a third genotype encoding based on GP trees: the representation of a secondary construction. The phenotype is slightly changed as we are not dealing with the truth table of a single Boolean function here, but rather with a family of functions. Following the same approach proposed in~\cite{10.1145/3512290.3528871}, a secondary construction takes as input a function of $n$ variables and gives a new function of $n+2$ variables as an output. From the point of view of the encoding, the leaves of the GP tree this time represent either four distinct \emph{seed} functions of $n$ variables, or the independent additional variables $x_{n+1}, x_{n+2}$. Thus, each tree always has six types of leaf nodes. The internal nodes are, instead, Boolean operators that take the input from the child nodes and forward the output to the parent node, with the root node representing the output of the construction. Figure~\ref{fig:gp-sec-constr} reports an example of how the GP tree is used to define a secondary construction. 

\subsubsection{Floating-point Encoding}
The last type of genotype representation that we considered in our experiments is the floating-point encoding, which is defined as a vector of continuous values. The genotype-to-phenotype mapping thus requires translating a sequence of floating-point numbers into a complete truth table with binary values. The idea behind this translation is that each value of the floating-point genotype represents a substring of bits in the genotype. To simplify the representation, all values in the floating-point vector are restricted to the interval $[0, 1]$. If the binary genotype size is denoted as $gsize$, the number of bits that a single coordinate of the floating-point vector represents can vary according to the following equation:
\begin{equation}
\label{eq:decode}
dec = \frac{gsize}{dim},
\end{equation}
where the parameter $dim$ denotes the size of the floating-point vector, and it can be modified if the genotype size is divisible by its value.

The first step of the translation procedure consists of converting each floating-point value in the genotype vector to an integer value. Since each floating-point value represents $dec$ bits as per Eq.~\eqref{eq:decode}, the size of the interval that maps to the same integer value is defined as:
\begin{equation}
\label{eq:intsize}
intsize = \frac{1}{dec}.
\end{equation}
Next, to obtain a distinct integer value for a given real number, every coordinate $d_i$ of the floating-point vector is divided by the interval size given by Eq.~\eqref{eq:intsize}, generating a sequence of integer values:
\begin{equation}
int\_value_i = \left \lfloor{\frac{d_i}{intsize}} \right\rfloor .
\end{equation}
Finally, the last step of the translation process involves mapping the obtained integer values into a binary string that can be used to construct the truth table of the individuals. For this purpose, we tested a simple binary encoding of the integer values. As an example, consider a genotype of $8$ bits. Suppose we want to represent it with $4$ real values; in this case, each real value encodes $2$ bits from the truth table. A string of two bits may have $4$ distinct combinations. Therefore, a single real value must be decoded into an integer value from $0$ to $3$. Since each real value is constrained to $[0, 1]$, the corresponding integer value is obtained by dividing the real value by $2^{-2} = 0.25$ and truncating it to the nearest smaller integer. Finally, the integer values are translated into the sequence of bits they encode. Following the above example, a floating-point vector like [0.71; 0.93; 0.13; 0.48] would be mapped into the integer vector [2; 3; 0; 1], which translates into the truth table ``10110001'' using a standard binary encoding. For further details on floating-point representation of Boolean functions, see~\cite{CarletDGJMP24}.

\subsection{Fitness Function}
\label{subsec:fitness}

As we stated in Section~\ref{sec:intro}, the cryptographic property that we aim to investigate in this paper is nonlinearity, i.e., the minimum distance of a Boolean function $f:\F_2^n \to \F_2$ from the set of all linear functions. Therefore, one simple choice for this task is to directly define the nonlinearity as the fitness function to be optimized by our evolutionary algorithms. However, one can come up with several other formulations, regardless of the underlying genotype representation and search algorithm employed. In this paper, we selected a fitness function based on the study of common choices in related works~\cite{DjurasevicJMP23}, and our previous experience on similar problems. More variants are, of course, possible, but they commonly include additional weight factors, making the tuning phase more complex.

The main problem encountered when only maximizing the nonlinearity value is that only the extreme values of the Walsh-Hadamard spectrum of a function are considered. Therefore, the information carried by the whole spectrum is not exploited to drive the evolution process. This choice makes the corresponding fitness landscape less smooth to explore, since an evolutionary algorithm has to potentially modify several genes at once in a candidate solution to ensure that the maximum absolute value of the spectrum decreases. For this reason, the fitness used in our experiments considers the whole Walsh-Hadamard spectrum of a Boolean function. In particular, we count the number of occurrences of the maximal absolute value in the spectrum, denoted as $\#max\_values$.
As higher nonlinearity corresponds to a lower maximal absolute value, the optimization objective is to minimize the occurrences of such a maximal value as much as possible. The underlying assumption here is that by having fewer occurrences of the maximal value in the spectrum, it becomes easier for the algorithm to reach the next nonlinearity value. In this way, we provide the algorithm with additional information, making the objective space more gradual. Formally, the fitness function is defined as:
\begin{equation}
\label{eq:second}
fitness = nl_{f} + \frac{2^n - \#max\_values}{2^n}.
\end{equation}
Remark that the second term never reaches the value of $1$: this case corresponds to the situation where we effectively reach the next nonlinearity level.

The function defined in Eq.~\eqref{eq:second} can be used to evaluate the fitness of a single function, which applies to most of the encodings described in the previous section. However, the secondary construction encoding does not produce a single function as a phenotype; rather, it is a family of functions obtained by evaluating a GP tree over a set of possible seed functions plus two independent additional variables. Thus, only for this specific encoding, we assess a GP tree that encodes a secondary construction by evaluating it on 10 groups of $n$-variable seed functions. The fitness is then the average fitness of the resulting $(n+2)$-variable functions, computed according to Eq.~\eqref{eq:second}.

\subsection{Algorithms Parameters}
\label{subsec:alg-param}
We now describe the remaining parameters of the evolutionary algorithms considered in our systematic evaluation that are specific to each of the adopted encodings, summarizing the generic ones at the end. 

\subsubsection{Bitstring Encoding}
Concerning the bitstring encoding, the variation operators that we employ are the simple bit mutation, which flips a randomly selected bit, and the shuffle mutation, which randomly shuffles the bits within a randomly selected substring. For crossover, we employ the one-point and uniform crossover operators. Given two parent bitstrings, the one-point crossover first selects a random crossover point. Then, a first offspring bitstring inherits the substring from the first position up to the crossover point from the first parent, and the substring from the crossover point to the last position from the second parent. A second offspring bitstring is created by swapping the inheritance order (first substring from the second parent, and second one from the first parent). On the other hand, the uniform crossover operator creates an offspring bitstring by going from left to right, and randomly copying at each position $i$ either the bit from the first or the second parent. Each time the evolutionary algorithm invokes a crossover or mutation operation, one of the previously described operators is randomly selected with uniform probability.

\subsubsection{Symbolic Encoding}
For the symbolic encoding, we employed the following set of Boolean operators for the internal nodes of the trees: OR, XOR, AND, AND2, XNOR, IF, and NOT. The operator AND2 is defined as $x_1 AND (NOT x_2)$, i.e., it corresponds to the usual AND gate but with the second input complemented. 
The operator IF is ternary, and given $x_1,x_2,x_3 \in \F_2$ it is defined as follows: if $x_1 = 1$ then it returns the second argument $x_2$ in output, and $x_3$ otherwise. As remarked in other related works, this function set is common when dealing with the evolution of Boolean functions with cryptographic properties~\cite{DjurasevicJMP23,CarletDGJMP24}. Notice that the same set of operators is used both for the direct search approach (i.e., when the GP tree is used to synthesize the truth table of a function directly) and in the secondary construction approach. In the latter case, when a seed function is used as an input for an internal node, the output will be a new Boolean function, possibly defined on a large number of variables, if the other inputs include an additional independent variable.

The genetic operators used in our experiments with tree-based GP are simple tree crossover, uniform crossover, size fair, one-point, and context preserving crossover~\cite{poli08:fieldguide} (selected uniformly at random), and subtree mutation. The idea of employing multiple genetic operators was based on the evidence gathered from preliminary experiments.

\subsubsection{Floating-point Encoding}
For the Floating-point (FP) genotype encoding, the number of bits that a single FP value represents can vary according to Eq.~\eqref{eq:decode}. Following the settings employed in related work~\cite{CarletDGJMP24}, all FP-based algorithms use the same setting with $dec=3$. Thus, a single value in the genotype encodes three bits. The interesting aspect of the floating-point representation is its versatility, since it can be used with any continuous optimization algorithm. In our experiments, we considered the following algorithms: Artificial Bee Colony (ABC)~\cite{karaboga2014comprehensive}, Clonal Selection Algorithm (CLONALG)~\cite{brownlee2007clonal}, CMA-ES~\cite{hansen2003reducing}, Differential Evolution (DE)~\cite{pant2020differential}, Optimization Immune Algorithm (OPTIA)~\cite{cutello2006real}, and a GA-based algorithm with floating-point chromosomes.

\subsubsection{Common Parameters and Settings}
We employ the same type of breeding strategy for the evolutionary algorithms used with the bitstring, symbolic, and floating-point encoding: a steady-state selection with a 3-tournament elimination operator (denoted SST). In each iteration of the algorithm, three individuals are chosen at random from the population for the tournament, and the worst one in terms of fitness value is eliminated. The two remaining individuals in the tournament are used with the crossover operator to generate a new child individual, which then undergoes mutation with probability $p_{mut} = 0.5$. Finally, the mutated child replaces the eliminated individual in the population. Preliminary tuning tests indicated that this combination of parameters exhibited the best performance, so we did not experiment further with other tournament sizes or mutation probabilities.

Regarding the remaining parameters of all other search algorithms included in our evaluation, we adopted the default values set in the ECF software framework.\footnote{\url{http://solve.fer.hr/ECF/}.} We considered the spaces of Boolean functions from $n=7$ to $n=13$ variables as problem instances. Finally, the termination condition is set at $10^6$ fitness evaluations for all encodings and instances. Each experiment is then repeated for 30 independent runs in order to obtain statistically sound results.

To determine whether a statistically significant difference between the results obtained by a group of different methods over the same problem instance exists, we carried out a statistical test in two stages: first, we applied the Kruskal-Wallis test with a significance value $\alpha=0.05$. If the $p$-value indicated significant differences, we further performed a pairwise Mann-Whitney U Test between all methods with the same $\alpha=0.05$, using Bonferroni correction since we considered multiple comparisons. In particular, the null hypothesis for both tests was that the compared samples were drawn from the same distribution. 
To summarize, we report in Table~\ref{tbl:list} the list of algorithms and encodings considered in our evaluation, together with their shorthand forms used in the next section.

\begin{table}
	\scriptsize
	\centering
	\caption{List of algorithms and encodings used in our experimental evaluation.}
	\label{tbl:list}
	\begin{tabular}{llll}
		\toprule
		Encoding & Description & Algorithm & Description \\
		\midrule
		\multirow{7}{*}{FP} & \multirow{7}{*}{Floating-point} & FP/ABC & Artificial Bee Colony \\
		                    &                                 & FP/CLONALG & Clonal Selection Algorithm \\
		                    &                                 & FP/CMAES & CMA-ES \\
		                    &                                 & FP/DE    & Differential Evolution \\
		                    &                                  & FP/OPTIA & Immune Optimization Algorithm \\
		                    &                                  & FP/SST   & Continuous Genetic Algorithm \\
		\midrule
		\multirow{2}{*}{GP} & \multirow{2}{*}{Symbolic} & GP/SST & GP -- direct search \\
		                    &                                    & GP/SCND & GP -- secondary construction \\
		\midrule
		TT                  & Bitstring & TT/SST & Genetic Algorithm \\
		\bottomrule 
	\end{tabular}
\end{table}

\section{Experimental Evaluation}
\label{sec:results}

This section presents the results obtained in our experiments by evaluating the various types of encodings and optimization algorithms described in the previous section. We start with a comparison of the basic optimization algorithms on the three encodings. Then, we consider the addition of a local search step and the restriction of the search space to rotation symmetric Boolean functions.

\subsection{Basic Optimization Algorithms}

\begin{figure}
    \centering
    \includegraphics[width=\textwidth]{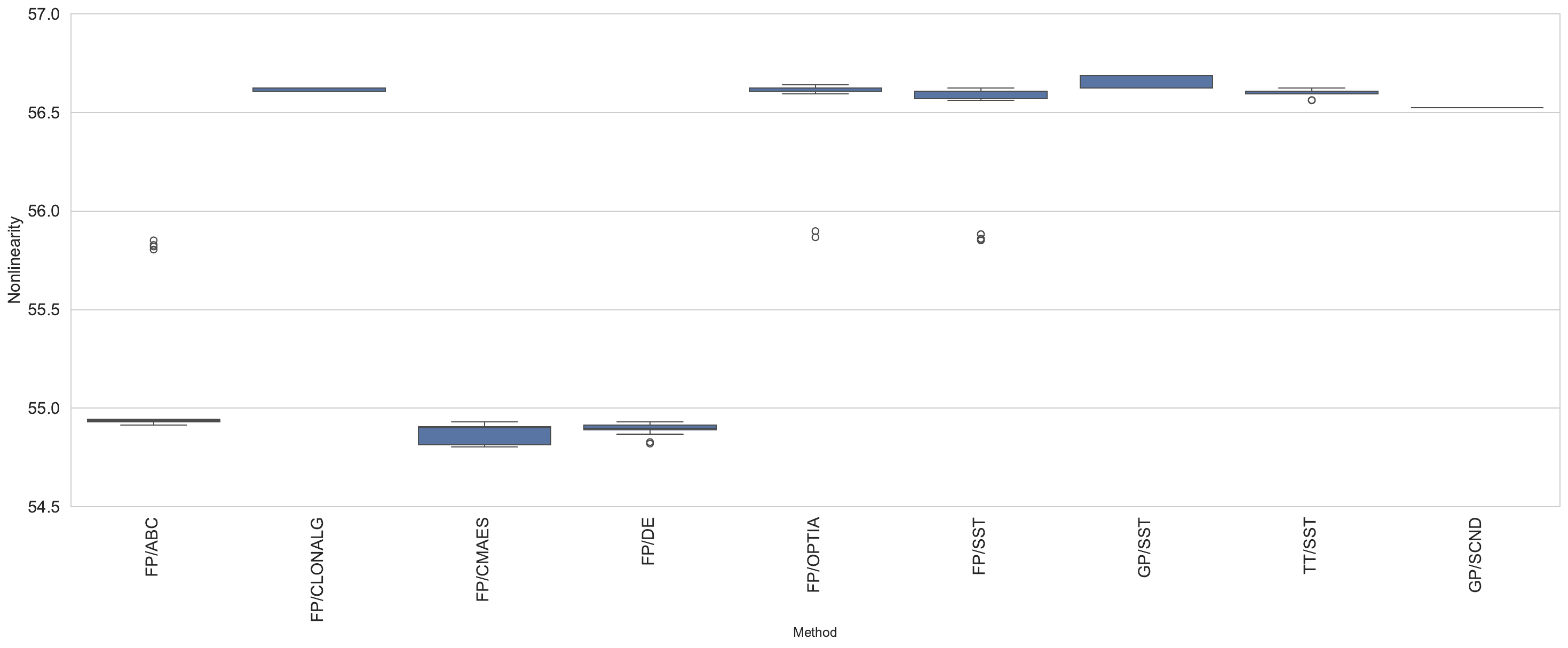}
    \caption{Distribution of the best fitness value for size $n=7$.}
    \label{fig:size7}
\end{figure}

\begin{figure}
    \centering
    \includegraphics[width=\textwidth]{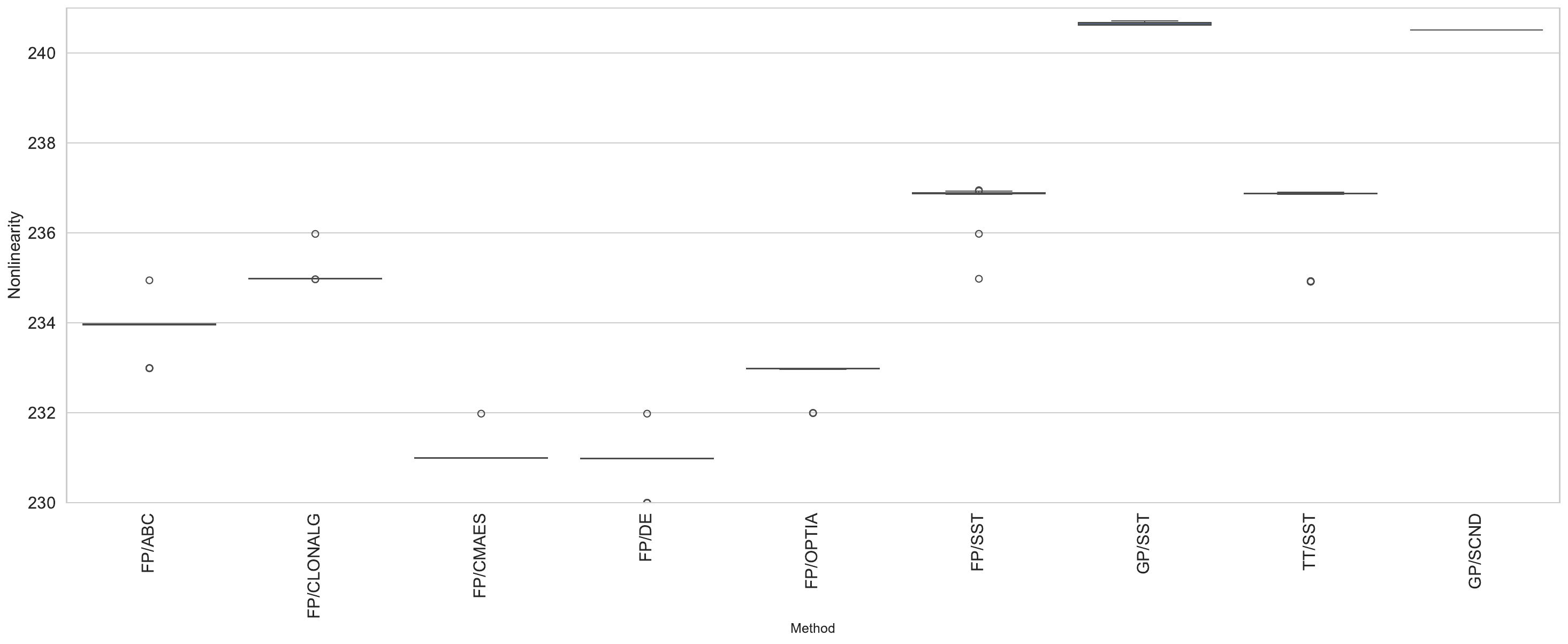}
    \caption{Distribution of the best fitness value for size $n=9$.}
    \label{fig:size9}
\end{figure}

\begin{figure}
    \centering
    \includegraphics[width=\textwidth]{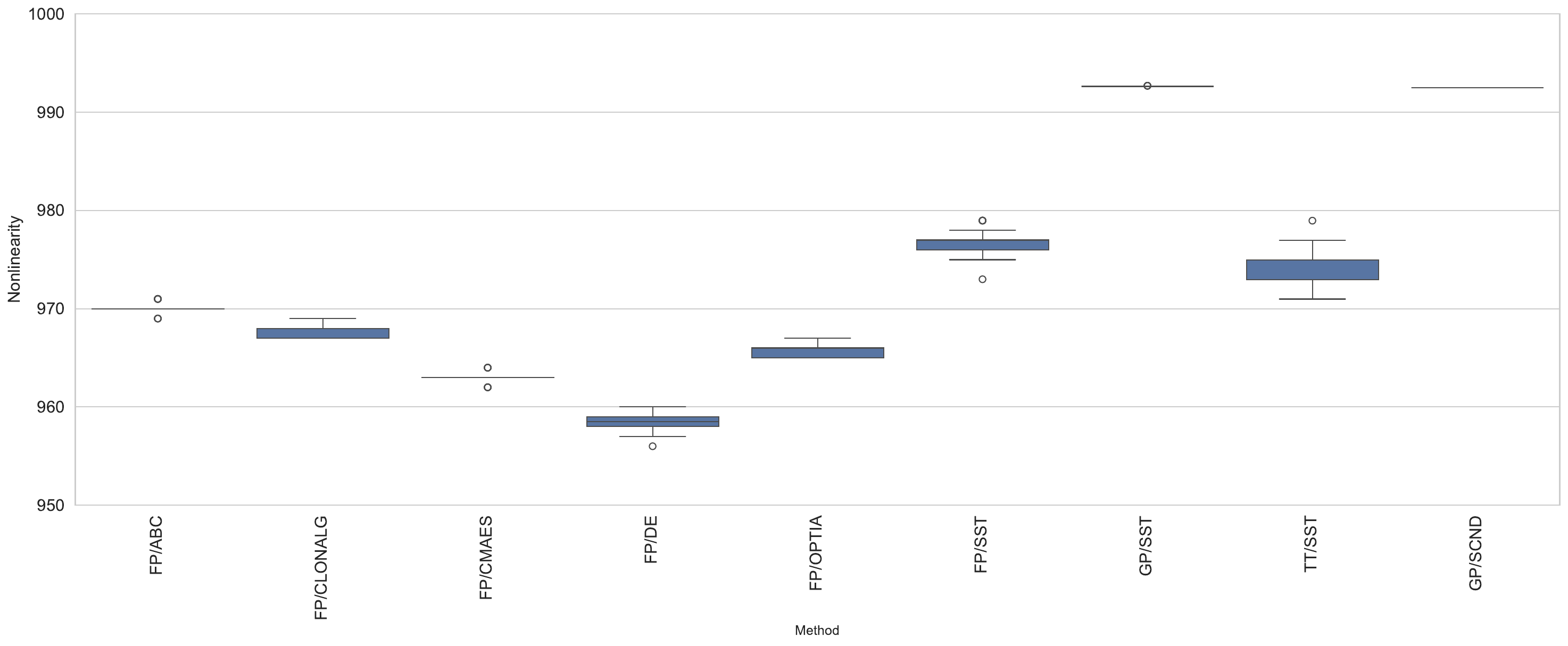}
    \caption{Distribution of the best fitness value for size $n=11$.}
    \label{fig:size11}
\end{figure}

\begin{figure}
    \centering
    \includegraphics[width=\textwidth]{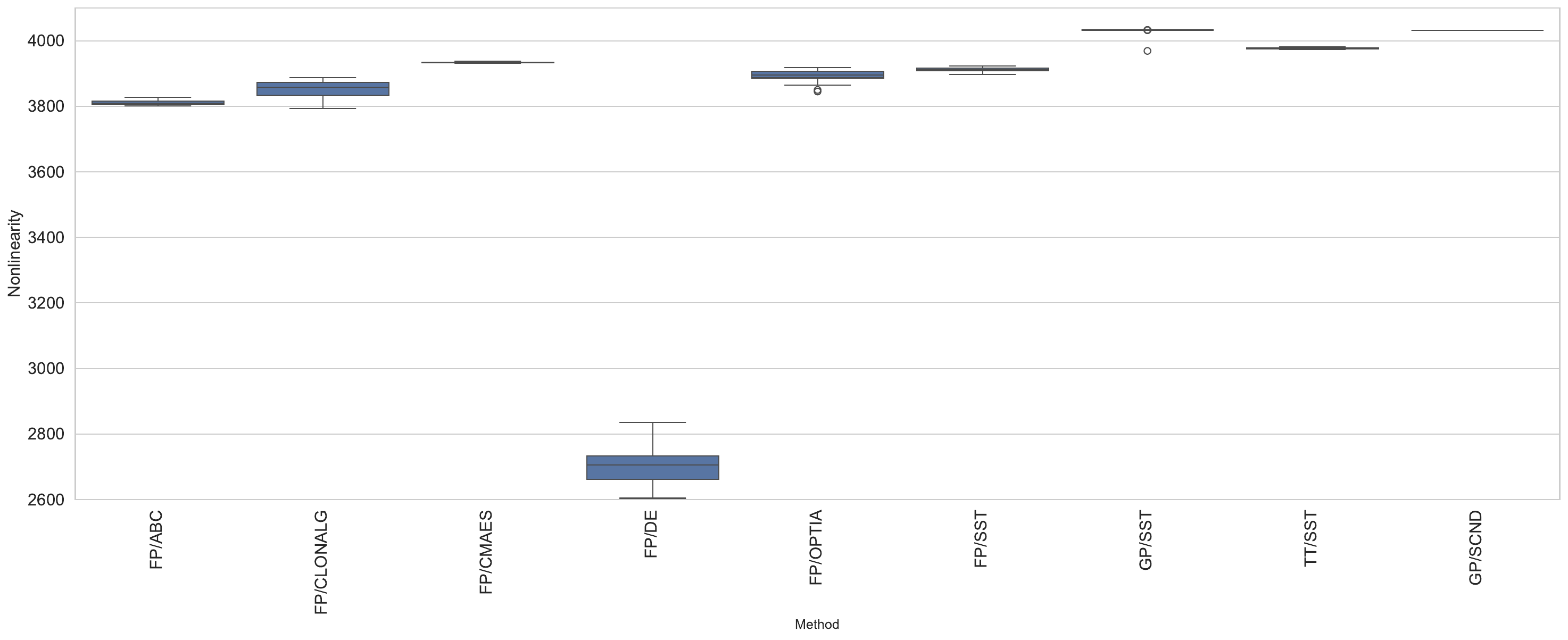}
    \caption{Distribution of the best fitness value for size $n=13$.}
    \label{fig:size13}
\end{figure}

The results for the combination of the three encodings, four problem instances, and nine optimization algorithms considered in our evaluation are summarized in Table~\ref{tbl:res}. In particular, each entry reports the maximum, average, and standard deviation of the best fitness value obtained over the 30 independent runs of the corresponding experiment. Further, the best average values are highlighted in bold for each problem instance. The distributions of the best fitness values achieved by each optimization method are also plotted in Figures~\ref{fig:size7} to~\ref{fig:size13}, for each problem instance.

\begin{table*}
\scriptsize
\centering
\caption{Summary of the results of the various representations and optimization algorithms (obtained fitness values).}
\label{tbl:res}
\begin{tabular}{@{}llccc|ccc@{}}
\toprule
\multirow{2}{*}{Enc.} & \multirow{2}{*}{Algorithm} & \multicolumn{3}{c}{7} & \multicolumn{3}{c}{9} \\ \cmidrule(l){3-8}
                      &                            & max.   & avg.  & std. & max.   & avg.   & std. \\ \midrule
\multirow{7}{*}{FP}             & ABC                        & 55.85  & 55.05 & 0.31 & 234.95 & 233.84 & 0.42 \\
                                & CLONALG                    & 56.63  & 56.62 & 0.01 & 235.98 & 235.01 & 0.18 \\
                                & CMAES                      & 54.93  & 54.65 & 0.50 & 231.98 & 231.02 & 0.18 \\
                                & DE                         & 54.93  & 54.90 & 0.03 & 231.98 & 230.79 & 0.48 \\
                                & OPTIA                      & 56.64  & 56.57 & 0.19 & 232.99 & 232.85 & 0.34 \\
                                & SST                        & 56.63  & 56.46 & 0.30 & 236.95 & 236.80 & 0.38 \\ \cmidrule(l){1-8}
\multirow{2}{*}{GP}                              & SST                        & 56.69  & \textbf{56.64} & 0.03 & 240.72 & \textbf{240.64} & 0.03       \\
                                                 & SCND                       & 56.53  & 56.53          & 0.00 & 240.51  & 240.51 & 0.00 \\ \cmidrule(l){1-8}
TT                              & SST                        & 56.63  & 56.60 & 0.02 & 236.91 & 236.55 & 0.74 \\ \midrule
\multicolumn{2}{@{}l}{\textbf{(continued)}} & \multicolumn{3}{c}{11} & \multicolumn{3}{c}{13} \\ \cmidrule(l){3-8}
Enc. & Algorithm & max.   & avg.  & std. & max.   & avg.   & std. \\ \midrule
\multirow{7}{*}{FP}             & ABC                        & 971.00 & 970.09 & 0.61 & 3827.00 & 3810.97 & 6.71 \\
                                & CLONALG                    & 969.00 & 967.76 & 0.57 & 3888.00 & 3853.40 & 24.37 \\
                                & CMAES                      & 964.00 & 963.00 & 0.52 & 3938.00 & 3934.23 & 1.41 \\
                                & DE                         & 960.00 & 958.50 & 1.01 & 2836.00 & 2701.12 & 58.23 \\
                                & OPTIA                      & 967.00 & 965.43 & 0.57 & 3918.00 & 3894.17 & 18.22 \\
                                & SST                        & 978.97 & 976.78 & 1.42 & 3923.00 & 3911.70 & 7.20 \\ \cmidrule(l){1-8}
\multirow{2}{*}{GP}                              & SST                        & 992.69 & \textbf{992.63} & 0.02 & 4032.69 & \textbf{4030.52} & 11.62 \\
                                                 & SCND                       & 992.51                   & 992.51                   & 0.00                     & 4032.55                  & 4032.55                  & 0.00 \\ \cmidrule(l){1-8}
TT                              & SST                        & 978.96 & 974.44 & 1.88 & 3980.99 & 3977.22 & 2.51 \\ \bottomrule
\end{tabular}

\end{table*}

Overall, the table and the plots clearly show that GP scores the best results across all four problem instances. Specifically, the difference in performance between GP and the other methods becomes more evident as the size of the problem increases. This trend is mostly independent of the underlying search strategy, either direct search or the evolution of secondary constructions. An additional benefit of GP with respect to the other methods and encoding is that it achieved a very small standard deviation value, which, in several cases, is the smallest among all the methods. Thus, the GP results are not dispersed, and the algorithm is rather stable. This is particularly evident for the secondary construction variant, which always scored a null standard deviation for all problem instances, meaning that the best solution evolved by GP always produces Boolean functions with the same nonlinearity value.

Concerning the TT and FP encoding, there is no method that is consistently better than the others, depending on the specific problem instance. In particular, the choice of the specific optimization algorithm had a significant impact on the performance when considering the FP encoding. Also, in this case, there is no single method that consistently scored the best results across all four problem sizes. However, the continuous genetic algorithm (FP/SST) seems to be the floating-point heuristic that achieved the best results overall.
 
Regarding the comparison with the best-known solutions, for size 7, each representation obtained the best-known nonlinearity of 56. We attribute this finding to the relatively small size of the search space for $7$-variable Boolean functions. Indeed, for larger problem instances, the obtained solutions are worse than the best-known ones for those numbers of variables (see Table~\ref{tab:nl}). It can be noticed, however, that the best methods achieved the maximum nonlinearity allowed by the quadratic bound.

The empirical findings discussed above were also mirrored by the statistical tests. Indeed, the Kruskal-Wallis test with $\alpha = 0.05$ yielded a $p$-value of 0 over all four problem instances. The Mann-Whitney U Test was then applied to investigate all pairwise comparisons of the methods under investigation. For $n=7$ variables, the results demonstrate that although GP/SST achieves the best results, it does not perform significantly better than the FP encoding with the CLONALG or OPTIA methods.  
Instead, the TT representation was significantly worse than GP but not significantly different from FP for some algorithms (namely, again CLONALG and OPTIA). Further, GP with the secondary construction search strategy performed worse than GP/SST, FP/CLONALG, and FP/OPTIA. We hypothesize that this effect is due to the small search space for the seed functions, since for $n=7$ they are defined over $5$ variables. Likely, the number of ways to combine highly nonlinear $5$-variable functions is not really large, making the problem more difficult for GP to come up with a suitable secondary construction.

For the remaining problem instances of $n=9$, $n=11$, and $n=13$, GP/SST achieves significantly better results than all other methods. In particular, even if the boxplots are barely distinguishable in Figures~\ref{fig:size9},~\ref{fig:size11}, and~\ref{fig:size13}, GP/SST scores a significantly higher fitness than GP/SCND. Furthermore, in these three problem instances, there is no significant difference between the TT and FP encodings when considering the result obtained by the best algorithm. Additionally, for $n=13$, TT/SST achieves equally good results as CMA-ES under the FP encoding.

Based on the previously outlined observations and analyses, we can conclude that GP with the direct search approach (GP/SST) is the most appropriate optimization algorithm for optimizing highly nonlinear Boolean functions, since it consistently scored the best results with respect to the other considered methods. Unfortunately, even these best results achieved by GP/SST fall short of the results reached by custom heuristics already published in the literature.

\subsection{Enhancements with Local Search and Rotation Symmetry}

Since the problem instance of $n=9$ variables was the smallest size where the optimization algorithms evaluated in the previous section did not achieve the best-known nonlinearity value of $241$, we considered two approaches to improve their performance: adding a local search step, and restricting the search space to rotation symmetric Boolean functions only. More precisely, we focused only on the genetic algorithm with the bitstring encoding (TT/SST) and the direct search approach with Genetic Programming (GP/SST). The reason is that these two methods obtained the best overall results as discussed in the previous sections. Moreover, even if GP/SCND proved to be a well-performing heuristic over all problem instances, the secondary construction approach is not straightforwardly amenable to the local search step, as well as to the restriction on rotation symmetric functions.

We implemented the local search step with two different strategies. In the first one, a mutation-based local search operator is applied on the genotype encoding of a Boolean function. In particular, the operator acts on a single solution and performs a number of random mutations. If a better solution is found after one of these mutations, the new solution replaces the current one, and the operator is applied again. The termination of this procedure occurs after a predefined number of mutations has been reached. The operator is applied after each generation, and acts on the current best solution in the population and on a number of other randomly selected solutions. In our experiments, the number of solutions undergoing local search was set to 5\% of the population size, and the number of trials (random mutations per individual) was set to 25. This operator is general, as it can be applied to any encoding.

The second local search strategy is instead specific to the bitstring encoding and performs individual bit flips instead of random mutations.
The difference with the first strategy, when applied to the bitstring encoding, is that this second strategy works in an exhaustive way. In particular, this local search operator performs all possible bit flips of the current solution, and terminates only if there is no improvement (i.e., a local optimum has been reached).

As mentioned above, we applied these two local search operators only to TT/SST and GP/SST as the most efficient variants; in the TT/SST case, three combinations were tested, with either the mutation (denoted as "-LS1") or bit flip operator (denoted as "-LS2"), or both (denoted as "-LS3"). The results with these modifications were not encouraging, despite the fact that our extended experimental design for this setting included $1\,000$ runs for every combination with a time limit of 2\,000 seconds, which corresponds to approximately 300 million evaluations per run. Further, the performance of GP was not affected, as it always found the same nonlinearity value of 240 in every run, with or without local search. The results for TT/SST slightly improved instead, and are reported in Table~\ref{tbl:ls}.

\begin{table}[t]
	\centering
	\caption{Results for the additional runs with the TT/SST and GP/SST algorithms and LS operators, $n=9$.}
	\label{tbl:ls}
	\begin{tabular}{@{}lccc@{}}
		\toprule
		& max.        & avg.   & std. \\ 
		\midrule
		TT/SST         & 236.91 & 236.55 & 0.74 \\
		TT/SST-LS1     & 238.83 & 237.98 & 0.95 \\
		TT/SST-LS2     & 238.87 & 238.58 & 0.64 \\
		TT/SST-LS3     & 238.87 & 238.69 & 0.48 \\ \midrule
		TT/SST-RI-LS1  & \textbf{241.75} & 240.75 & 0.05 \\
		TT/SST-RI-LS2  & 240.88 & 240.80 & 0.03 \\
		TT/SST-RI-LS3  & 240.90 & 240.79 & 0.04 \\ 
		\midrule
		GP/SST         & 240.72 & 240.64 & 0.03 \\
		GP/SST-PART    & 240.62 & 238.99 & 0.84 \\
		GP/SST-PART-LS & 240.64 & 239.29 & 0.98 \\
		GP/SST-FULL    & 240.63 & 239.03 & 0.74 \\
		GP/SST-FULL-LS & 240.61 & 238.90 & 0.85 \\
		\bottomrule
	\end{tabular}
\end{table}

After the experiments with the local search variants, we also considered a second type of enhancement, namely the restriction of the search space to rotation symmetric Boolean functions. This constraint can be achieved by using the ad-hoc genotypes described in Section~\ref{subsec:sol-enc} under the bitstring and symbolic encodings. In particular, in the $n=9$ problem instance under consideration, the bitstring encoding genotype for rotation symmetric functions consists of only 60 bits, as opposed to 512 in the generic space of all $9$-variable Boolean functions. These results are also included in Table~\ref{tbl:ls} and denoted with the suffix "-RI" (Rotation Invariant) in the combinations related to TT/SST. For GP, we used instead the two approaches to restrict the synthesis of the phenotype to rotation symmetric functions starting from the syntactic trees. In the first approach (GP/PART), we evolved trees for Boolean functions of $n=6$ variables, since the corresponding truth tables have size $2^6=64$ bits, which is larger than the $60$ required to specify a $9$-variable rotation symmetric function. In the second approach, we evolved trees of $9$-variable functions, but evaluated them only on the representatives of the $60$ rotation classes, filling the remaining inputs of the truth table accordingly. In both cases, we also considered the variant where the LS operator is applied, respectively denoted as GP/PART-LS and GP/FULL-LS.

One can see from Table~\ref{tbl:ls} that enforcing rotation symmetry leads to improvements in the results of the GA with the bitstring encoding (variants TT/SST-RI). This variant also achieved the maximum nonlinearity of 241 for the TT-RI-LS1 combination. 

The bottom part of Table~\ref{tbl:ls} reports the descriptive statistics for the considered GP combinations with rotation symmetry and local search. It can be noticed that constraining the search space to rotation symmetric functions does not improve the maximum fitness value on GP, whether the LS step is applied or not. More surprisingly, one can see that the average performance of GP slightly drops on rotation symmetric functions, with nonlinearity values of around 238 or 239, instead of 240 as in the original version of GP. 

\begin{figure}[t]
	\centering
	\includegraphics[width=\textwidth]{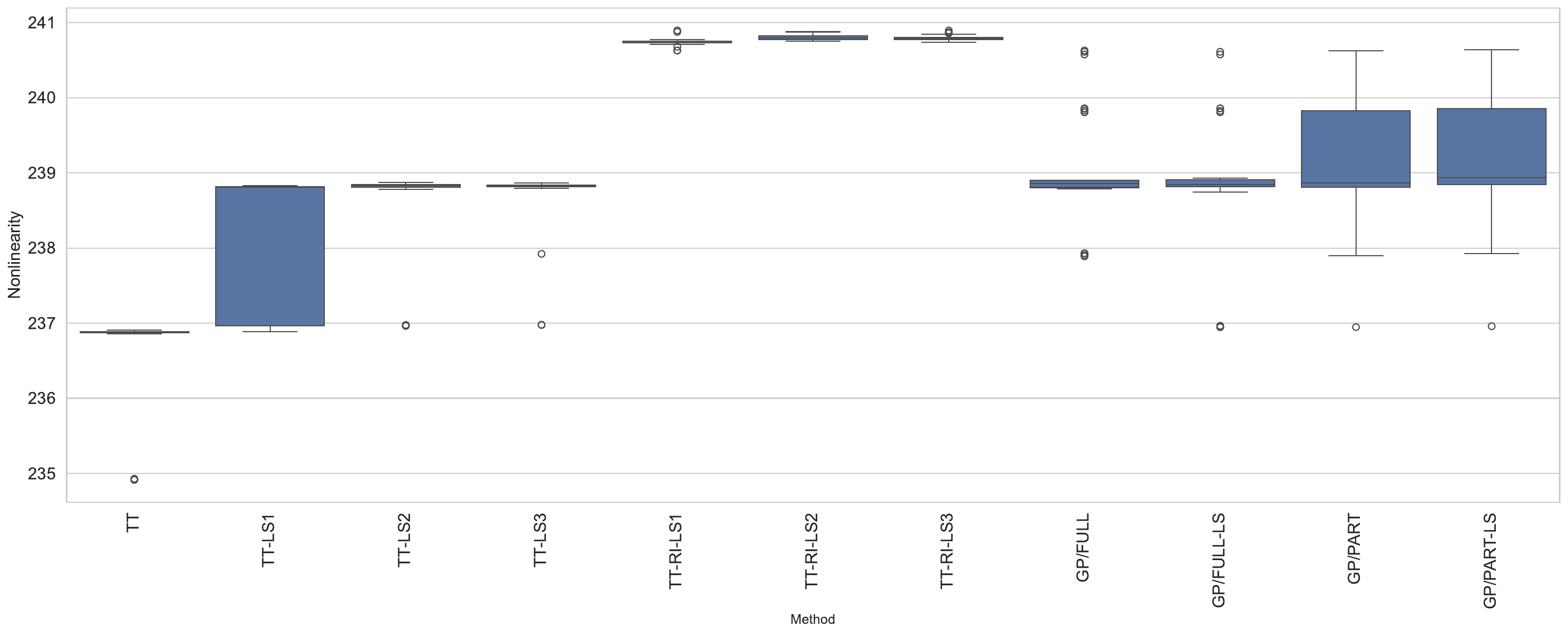}
	\caption{Boxplot representation of the results for the application of LS operators on the TT representation.}
	\label{fig:boxls}
\end{figure}

To better outline the effect of the different LS strategies on the results, Figure~\ref{fig:boxls} provides the boxplots for the distributions of the best fitness value. It can be observed that the application of the local search operators positively affects the results under the TT encoding, especially the bit flip operator. On the other hand, the observation made on Table~\ref{tbl:ls} about the hindrance that LS introduces in GP is also corroborated by the corresponding boxplots in Figure~\ref{fig:boxls}.

Concerning the statistical analysis of these distributions, the Kruskal-Wallis test yielded again a $p$-value of $0$, thereby indicating that this group of algorithms' variants perform significantly differently. The second stage of the analysis, through the Mann-Whitney U test, demonstrates that, by using LS, it is possible to improve the results significantly compared to the basic algorithm for the GA with bitstring encoding, but not for GP. Regarding the different LS operators, the statistical analysis indicates that there is no significant difference between them. 

\subsection{Discussion}
\label{subsec:disc}

We conclude our evaluation by discussing some insights emerging from the results presented in the previous sections.
Concerning GP, it is remarkable that the basic version with the direct search approach is the overall best-performing algorithm, while the variants that enforce rotation symmetry or apply a local search operator achieve subpar results. This finding is particularly interesting, as it suggests that constraining the search space to specific functions or exploiting the search process with local search seems to benefit only genetic algorithms with a bitstring encoding. Indeed, it might be the case that acting on a tree-based representation introduces more disruptive changes than what is actually needed to slightly improve the nonlinearity values. After all, GP already manages to find solutions with a nonlinearity close to the best-known one for nine inputs. Hence, applying random mutations to a near-optimal tree could be more detrimental than beneficial for a candidate solution in this context. A similar hypothesis could apply to the rotation symmetry constraint: the tree-based encoding might simply not be appropriate to evolve rotation symmetric Boolean functions with the encodings we explored in this paper. In particular, the issue could lie in the fact that GP is ``wasting'' genetic information, since the genotypes are larger than what is actually needed to define a rotation symmetric function. In the first case, we evolve a GP tree such that the corresponding Boolean function has a truth table which is larger than the size required to define the bitstring of rotation classes. In the second case, we evolve a GP tree that is only partially evaluated. This could imply that GP is evolving parts of candidate solutions that, in retrospect, are not useful for the optimization process.

We now consider how our best results obtained with GA under the bitstring encoding constrained to rotation symmetric functions compare with the custom heuristic approach developed in~\cite{4167738}. The authors of that work reported that, among 200 million RSBFs of nine variables evaluated with a steepest descent-like algorithm, five have nonlinearity 241. Clearly, this is more successful than what we achieved, as we found only one such function over 300 million fitness evaluations. Therefore, the question becomes where this difference in performance arises. We highlight three factors that might contribute to this difference. First, although Kavut et al. also considered the sum of square errors of the Walsh-Hadamard values, their overall objective function differs from our fitness function. It could thus be the case that our fitness function yields a more irregular landscape, making it more challenging for an EA to converge on a solution of nonlinearity 241. Second, their steepest descent-like algorithm introduced a step where, once the cost cannot be minimized further, backtracking is deterministically applied by applying a move that corresponds to the smallest possible cost increase. Finally, in our experiments, we did not use exclusively a local search algorithm, but rather we combined it as an additional step of an evolutionary algorithm. Based on the results that we obtained in our systematic evaluation, we conclude that local search is a crucial process, which indicates that EA operators are either 1) too disruptive or 2) reach local optima and cannot produce a small change required to improve the fitness value. As a matter of fact, increasing the nonlinearity from 240 to 241 requires---in the best case scenario---only a single change in the truth table representation. 

\section{Conclusions and Future Directions}
\label{sec:conclusions}

In this paper, we systematically evaluated the design of highly nonlinear Boolean functions defined over an odd number of variables via evolutionary algorithms. The experiments included three solution encodings, four problem instances, and nine different search algorithms. Overall, the results indicate that GP scores the best results, regardless of the fact that GP works with general Boolean functions. Unfortunately, even the best results obtained by GP fall short of the best-known results reached with custom heuristics, except for the smallest problem instance of $n=7$ variables. Further, we added several local search variants to our best EAs, i.e., GA with the bitstring encoding and GP. Those modifications did not help GP but improved the GA results. Moreover, one combination of the GA with bitstring encoding and local search operators even produced an example of a $9$-variable Boolean function with nonlinearity 241. Interestingly, the results suggest that adding local search or narrowing the space to rotation symmetric functions actually hamper the performance of GP.

There are several interesting avenues for future research on this problem. As a general research theme, the fitness landscape of the optimization problems related to Boolean functions is poorly understood. To our knowledge, there exists only one study that addressed it in the general case~\cite{JAKOBOVIC2021107327}. In this respect, two possible research directions would be to 1) leverage the insights obtained in that study, concerning, e.g., the initialization strategy, and apply them to our evaluation framework to see if the performances of our search algorithms are affected, and if so to what extent they are; and 2) perform a fitness landscape analysis for the space of rotation symmetric Boolean functions. This might help to better understand the dynamics of the optimization algorithms considered in this study over that particular search space. In principle, such an investigation could even help to shed light on why GA combined with local search was the only metaheuristic able to produce a Boolean function with nonlinearity 241.

Further, there are countless other variations that one could consider in our experimental setup: besides optimization algorithms based on a different metaphor, it could be interesting to explore other variants of GA and GP, for instance, by developing crossover and mutation operators that are more suitable for this type of problem. For instance, a possible direction here would be to adapt the balanced crossover operators investigated in~\cite{manzoni20} for GA to the setting of rotation symmetric Boolean functions. Further, one may also consider semantically-aware operators for GP, which can tweak the genotype trees by enforcing certain semantic constraints. In this regard, a concrete direction could be to adapt the semantic mutation operators developed by Husa et al.~\cite{DBLP:journals/gpem/HusaS24} for evolving bent functions to the case of odd-sized highly nonlinear Boolean functions. 

\bibliographystyle{abbrv}
\bibliography{bibliography}

\end{document}